\documentclass[english,11pt]{article}
\usepackage[T1]{fontenc}
\usepackage[latin9]{inputenc}
\usepackage{geometry}
\usepackage{amsmath}
\usepackage{setspace}
\makeatletter

\providecommand{\tabularnewline}{\\}

\newcommand{\lyxaddress}[1]{
\par {\raggedright #1
\vspace{1.4em}
\noindent\par}
}

\makeatletter \renewcommand\@biblabel[1]{#1. } \makeatother
\makeatletter 
\usepackage{url}
\usepackage{fancyhdr}
\usepackage{multirow}

\makeatother

\usepackage{babel}
\begin{document}

\title{Automatic skin lesion segmentation with fully convolutional-deconvolutional
networks}

\date{}
\maketitle
\thispagestyle{fancy} 

\addtolength{\headheight}{0.0mm} 

\renewcommand{\headrulewidth}{0pt}

\renewcommand{\headsep}{0mm} 

\centerline{Yading Yuan}

\lyxaddress{\centerline{Department of Radiation Oncology}
\centerline{Icahn School of Medicine at Mount Sinai}
\centerline{New York, NY 10029}
\centerline{yading.yuan@mssm.edu}}
\begin{abstract}
This paper summarizes our method and validation results for the ISBI
Challenge 2017 - Skin Lesion Analysis Towards Melanoma Detection -
Part 1: Lesion Segmentation. 
\end{abstract}

\section{INTRODUCTION}

Automatically segmenting melanoma from the surrounding skin is an
essential step in computerized analysis of dermoscopic images. However,
this task is not trivial because melanoma usually has a large variety
of appearance in size, shape, and color along with different types
of skin and texture. Meanwhile, some lesions have irregular and fuzzy
borders, and in some cases the contrast between lesion and the surrounding
skin is low. In addition, artifacts and intrinsic cutaneous features,
such as hairs, frames, blood vessels and air bubbles can make the
automatic segmentation more challenging. Researchers have developed
various computer algorithms to conquer these challenges, and a recent
survey can be found in \cite{key-1}.

We proposed a framework based on deep fully convolutional-deconvolutional
neural networks (CDNN) \cite{key-2,key-3} to automatically segment
skin lesions in dermoscopic images. Instead of developing sophisticated
pre- and post-processing algorithms and hand-crafted features, we
focus on designing appropriate network architecture and effective
training strategies such that our deep learning model can handle images
under various acquisition conditions. 

\section{MATERIALS AND METHODS}

\subsection{Database}

We participated the part I of ISBI Challenge 2017 - Skin Lesion Analysis
Towards Melanoma Detection: Lesion Segmentation. The training dataset
includes $2000$ dermoscopic images in .jpg format and the corresponding
lesion masks in .png format. The images are of various dimensions.
The lesion types involved include nevus, seborrhoeic keratosis and
malignant melanoma. The goal is to produce accurate binary masks of
various skin lesions against a variety of background. Besides training
set, the organizers provide a validation dataset that includes $150$
images. The participants can submit the binary masks of these $150$
images and evaluate the segmentation performance online. Additional
test dataset with $600$ images is provided for final evaluation.
The final rank is based on Jaccard index.

\subsection{Our Approach}

\subsubsection{Architecture}

We train a CDNN to map from input dermoscopic image to a posterior
probability map. The network contains $29$ layers with about $5$M
trainable parameters. Table \ref{tab:cdnn} describes the architectural
details. We fixed the stride as 1 and use Rectified Linear Units (ReLUs)
as the activation function for each convolutional/deconvolutional
layer. For output layer, we use sigmoid as the activation function.
Pixel-wise classification is performed and CDNN is essentially served
as a filter that projects the entire input image to a map where each
element represents the probability that the corresponding input pixel
belongs to the tumor. In order to address the conflict between multi-scale
information aggregation and full-resolution pixel-wise classification,
we implement a strategy of using upsampling and deconvolutional layers
to recover lost resolution while carrying over the global perspective
from pooling layers \cite{key-3}. Batch normalization is added to
the output of every convolutional/deconvolutional layer to reduce
the internal covariate shift.

\begin{table}
\centering

\caption{Architectural details of the proposed CDNN model (Abbrevations: conv:
convolutional layer; pool: max-pooling layer; decv: deconvolutional
layer, ups: upsampling layer). \label{tab:cdnn}}

\begin{tabular}{cccccc}
 &  &  &  &  & \tabularnewline
\hline 
\hline 
\noalign{\vskip2mm}
Conv & Filter size & No. of features & Deconv & Filter size & No. of features\tabularnewline[2mm]
\hline 
conv-1-1 & $3\times3$ & $16$ & decv-1 & $3\times3$ & $256$\tabularnewline
conv-1-2 & $3\times3$ & $32$ & ups-1 & $2\times2$ & $256$\tabularnewline
pool-1 & $2\times2$ & $32$ & decv-2-1 & $3\times3$ & $256$\tabularnewline
conv-2-1 & $3\times3$ & $64$ & decv-2-2 & $3\times3$ & $128$\tabularnewline
conv-2-2 & $3\times3$ & $64$ & ups-2 & $2\times2$ & $128$\tabularnewline
pool-2 & $2\times2$ & $64$ & decv-3-1 & $4\times4$ & $128$\tabularnewline
conv-3-1 & $3\times3$ & $128$ & decv-3-2 & $3\times3$ & $128$\tabularnewline
conv-3-2 & $4\times4$ & $128$ & ups-3 & $2\times2$ & $128$\tabularnewline
pool-3 & $2\times2$ & $128$ & decv-4-1 & $3\times3$ & $64$\tabularnewline
conv-4-1 & $3\times3$ & $256$ & decv-4-2 & $3\times3$ & $32$\tabularnewline
conv-4-2 & $3\times3$ & $256$ & ups-4 & $2\times2$ & $32$\tabularnewline
pool-4 & $2\times2$ & $256$ & decv-5-1 & $3\times3$ & $16$\tabularnewline
conv-5 & $3\times3$ & $512$ & output & $3\times3$ & $1$\tabularnewline
\hline 
\hline 
 &  &  &  &  & \tabularnewline
\end{tabular}
\end{table}

\subsubsection{Pre-processing}

A simple pre-processing is employed to facilitate the following learning
procedure while preserving the original image information. Besides
the original RGB channels, we also include the three channels in Hue-Saturation-Value
color space, as well as the L channel (lightness) in CIELAB space.
Each channel is rescaled to $[0,\,1]$. By observing most of images
in the training set have a height-to-width ratio of $3:4$, we resize
the images to $192\times256$. 

\subsubsection{Training}

We train the network using Adam optimization \cite{key-4}with batch
size of $16$ to adjust the learning rate based on the first and the
second-order moments of the gradient at each iteration. The initial
learning rate is set as $0.003$. In order to reduce overfitting,
we use dropout with $p=0.5$ before conv-4-1 and decv-5-1 in Table
\ref{tab:cdnn}, and employ two types of image augmentations to further
improve the robustness of the proposed model under a wide variety
of image acquisition conditions. One consists of a series of geometric
transformations, including randomly flipping, shifting, rotating as
well as scaling. The other type focuses on randomly normalizing the
contrast of each channels in the training images. Note that these
augmentations only require little extra computation, so the transformed
images are generated from the original images for every mini-batch
within each iteration.

We design a loss function based on Jaccard distance in this study:
\begin{equation}
L_{d_{J}}=1-\frac{\underset{i,j}{\sum}(t_{ij}p_{ij})}{\underset{i,j}{\sum}t_{ij}^{2}+\underset{i,j}{\sum}p_{ij}^{2}-\underset{i,j}{\sum}(t_{ij}p_{ij})},\label{eq:ja-loss}
\end{equation}
where $t_{ij}$ and $p_{ij}$ are target and the output of pixel $(i,\;j)$,
respectively. As compared to conventionally used cross-entropy, the
proposed loss function is directly related to image segmentation task
because Jaccard index is a common metric to assess medical image segmentation
accuracy, especially in this challenge. Meanwhile, this loss function
is well adapted to the problems with high imbalance between foreground
and background classes as it doesn't require any class re-balancing.

\subsubsection{Post-processing}

We use a dual-thresholds method to generate a binary tumor mask from
the CDNN output. A relatively high threshold ($th_{H}=0.8$) is firstly
applied to determine the tumor center, which is calculated as the
centroid of the region that has the largest mass among the candidates
from thresholding. Then a lower threshold $th_{L}=0.5$ is applied
to the output map. After filling small holes with morphological dilation,
the final tumor mask is determined as the region that embraces the
tumor center. Finally, a bagging-type ensemble strategy is implemented
to combine outputs of 6 CDNNs to further improve the image segmentation
performance on the testing images.

\subsubsection{Implementation}

Our method was implemented with Python based on Theano \cite{key-5}
and Lasagne\footnote{http://github.com/Lasagne/Lasagne}. The experiments
were conducted on a Dell XPS 8900 desktop with Intel (R) i7-6700 3.4
GHz GPU and a GPU of Nvidia GeForce GTX 1060 with 6GB GDDR5 memory.
The total number of epochs was set as $500$.

\section{RESULTS}

Our method yielded an average Jaccard index of $0.784$ on the online
validation dataset.

\end{document}